\documentclass{article}

\usepackage{arxiv}

\usepackage[utf8]{inputenc} 
\usepackage[T1]{fontenc}    
\usepackage{hyperref}       
\usepackage{url}            
\usepackage{booktabs}       
\usepackage{amsfonts}       
\usepackage{nicefrac}       
\usepackage{microtype}      
\usepackage{lipsum}
\usepackage{graphicx}
\usepackage{amsmath}
\usepackage[capitalise]{cleveref}
\crefname{equation}{Eq.}{Eqs.}
\Crefname{equation}{Eq.}{Eqs.}
\usepackage{microtype}      
\usepackage{mathtools}      
\graphicspath{ {./images/} }
\usepackage{float}
\usepackage[round]{natbib}
\usepackage{xcolor}

\title{Forecast skill is not decision skill: evidence from weather-dependent decision tasks}

\author{
Kornelius Raeth\textsuperscript{1}\thanks{This work was supported by the DFG -- EXC number 2064/1 -- Project number 390727645 and by the German Federal Ministry of Research, Technology and Space (BMFTR) through the FEAT project (grant number 16IS22073A).} \hspace{2em}
Nicole Ludwig\textsuperscript{1,2}\\[1ex]
\textsuperscript{1}University of Tübingen, Tübingen AI Center, Germany\\
\textsuperscript{2}University of Augsburg, Germany\\[1ex]
Correspondence to: \texttt{kornelius.raeth@uni-tuebingen.de}
}

\begin{document}
\maketitle
\begin{abstract}
Standard weather forecast evaluations focus on the forecaster's perspective and on a statistical assessment comparing forecasts and observations. In practice, however, forecasts are used to make decisions, so it seems natural to take the decision-maker's perspective and quantify the value of a forecast by its ability to improve decision-making. Decision calibration provides a novel framework for evaluating probabilistic forecast performance at the decision level rather than the forecast level. We evaluate decision calibration to compare a Machine Learning and a classical numerical weather prediction model on various weather-dependent decision tasks, though the framework is applicable to any set of forecast models. We find that model performance at the forecast level does not reliably translate to performance in downstream decision-making: some performance differences only become apparent at the decision level, and even among seemingly similar decision tasks, model rankings can change. Our results confirm that typical forecast evaluations are insufficient for selecting the optimal forecast model for a specific decision task.
\end{abstract}



\section{Introduction}
Weather forecasts have a daily impact on people's lives by providing early warnings of extreme events and guiding decision-making in critical industries like aviation, public health, agriculture, and energy. With climate change intensifying weather extremes, accurate forecasts are increasingly important. Recently, Machine Learning weather prediction (MLWP) models \citep{lam2023, lang2024, chen2023} are rivaling or surpassing the performance of classical numerical weather prediction (NWP) models across many metrics and variables.
However, deterministic point forecasts hide the uncertainty inherent in weather prediction due to the chaotic and stochastic nature of the atmosphere and modeling limitations, making uncertainty quantification essential for forecast validity \citep{murphy1993, palmer2019}.
To address this, ensemble forecasts were introduced into NWP systems \citep{molteni1996ecmwf}. Probabilistic MLWP models have recently achieved performance comparable to traditional numerical models \citep{price2023,lang2024b, couairon2024, bonev2025}. From a forecaster's perspective, a good probabilistic forecast should be maximally sharp while remaining calibrated \citep{gneiting2007}.
Sharp forecasts have concentrated predictive distributions, i.e., low predictive variance.
Calibration refers to the statistical consistency between forecasts and observations
(e.g., a $90\%$ prediction interval derived from the forecast should cover the observation $90\%$ of the time).
But are those really the key properties of a forecast that are relevant to end users who use forecasts to make decisions?

\paragraph{Good forecasts improve decision-making}
Ultimately, forecasts are used to make decisions, and their value lies in their ability to improve decision-making \citep{sweeney2020, zhu2002, murphy1994}.
Yet, typical forecast evaluations consider only the forecaster's perspective and ignore downstream decision-making tasks \citep{maciejowska2025}. 
In theory, forecasts fulfilling distribution calibration \citep{song2019} yield optimal downstream decisions. However, distribution calibration represents a strong notion of calibration that is difficult to attain and verify in practice \citep{zhao2021, sahoo2021}. While weaker forms of calibration can be achieved and validated, they typically offer no guarantees for improved decision-making.
Common diagnostic tools for calibration in weather forecasting include probability integral transform (PIT) histograms, proper scoring rules such as the continuous ranked probability score (CRPS) \citep{gneiting2007}, and spread skill ratios (SSR) \citep{fortin2014}. For a more detailed discussion of common calibration diagnostics, see Appendix \Cref{sec:appdx:cal}.
Since each metric captures a different aspect of forecast quality, disagreement between metrics on model rankings is common and expected.
More importantly, while all of these metrics quantify some form or aspect of calibration, they may omit forecast properties relevant to decision-making \citep{murphy1993}. This highlights the need to evaluate calibration from the decision-maker's perspective, at the level of decisions rather than forecasts.

\begin{figure}
    \centering
    \includegraphics[width=1.0\linewidth]{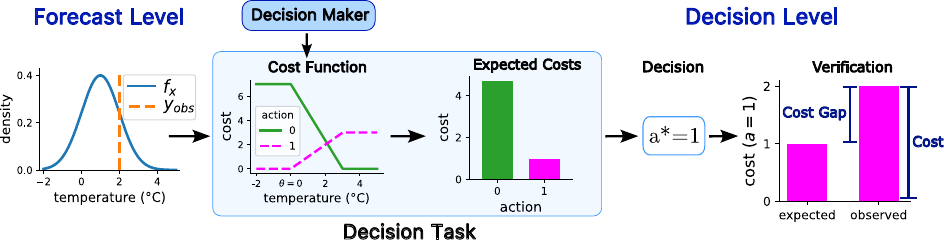}
    \caption{Summary diagram of our approach. Traditional forecast verification operates at the forecast level, comparing forecast $f_x$ directly to observation $y_{\text{obs}}$. Our approach inserts a decision task prior to evaluation, shifting verification to the decision level: the probabilistic forecast is combined with a cost function to select the action with minimal expected cost. Evaluation then considers both the observed cost $c(a=1, y_{\text{obs}})$ and the cost gap between observed and expected costs, jointly capturing decision quality and decision calibration.}
    \label{fig:overview}
\end{figure}

\paragraph{Calibration as precise cost estimation}
Decision tasks can be modeled via cost functions of actions and observations. These cost functions come from decision makers with their own costs and preferences. Forecasts are then used by the decision maker to estimate costs and make decisions, i.e., select the optimal action (see \Cref{fig:overview}). In this setup, calibration is framed as precise cost estimation \citep{derr2025}: a forecast is considered calibrated if it correctly anticipates the costs of a particular action (expected costs = observed costs). This notion is called decision calibration \citep{zhao2021} and shifts the focus of evaluation from distribution matching to only those forecast properties relevant to the specific decision task.

\paragraph{Contributions}
We observe a mismatch between current forecast verification practices and what decision makers care about in practice. We bridge this gap by presenting the first study to evaluate state-of-the-art weather prediction models at the decision level using the decision calibration framework (see \Cref{fig:overview}). 
\begin{itemize}
    \item We evaluate performance on decision tasks motivated by applications in core weather-dependent sectors: frost protection, heat protection, and wind power dispatch.
    \item We investigate to what extent calibration performance at the forecast level translates into performance in downstream decision-making.
    \item We compare state-of-the-art numerical and ML weather prediction models with respect to decision calibration.
\end{itemize}

\paragraph{Related work}
Existing work on decision calibration primarily focuses on medical applications (disease detection, hospital acceptance) \citep{zhao2021, sahoo2021}.
Studies on the value of weather forecasts investigate the relative economic value (REV) of ensemble forecasts \citep{zhu2002, richardson2000, mylne2002, price2023}, quantifying incurred costs using simple cost-loss ratio cost functions.
Our framework goes beyond REV in two fundamental ways: first, it additionally accounts for decision calibration — the degree to which anticipated costs match incurred costs, a dimension entirely absent from REV; second, it accommodates arbitrary cost functions and action spaces, recovering REV as the special case of binary threshold decisions with a cost-loss cost function.
Empirical research has studied weather-related decision-making, investigating how individual decision-makers, often non-experts, use probabilistic weather forecasts in practice \citep{joslyn2012, mohrlen2022}. Our framework takes a normative perspective and evaluates forecast systems assuming optimal decision-making.

\section{Decision-making and calibration}

\subsection{Setup and notation}
\label{notation}
We consider a regression task where the goal is to learn a mapping from input features $X \in \mathcal{X} \subset \mathbb{R}^m$ to targets $Y \in \mathcal{Y} \subset \mathbb{R}$, where $\mathcal{X}$ denotes the $m$-dimensional real-valued input space and $\mathcal{Y}$ the real-valued target space. $X$ and $Y$ are random variables distributed according to a data generating process $P_{X,Y}$. We are interested in probabilistic modeling, so given an input $X$, we write $F_{X}$ to denote the predictive cumulative distribution function (CDF) over targets $\mathcal{Y}$ that ideally recovers the true CDF of $Y|X$. 

\subsection{Optimal decision-making under uncertainty}
In the context of optimal decision-making, a decision maker with action space $\mathcal{A}$ selects the action that minimizes expected costs under the predictive distribution. To formalize this, consider a cost function $c: \mathcal{A} \times \mathcal{Y}$, that quantifies the cost $c(a,y)$ incurred by a decision maker when taking action $a \in \mathcal{A}$ and observing $y \in \mathcal{Y}$. Cost functions encode the costs and preferences of a particular decision maker and therefore vary across decision makers. Given a probabilistic prediction $F_x$ over $\mathcal{Y}$, the Bayes decision rule $\delta_c$ selects the action that yields the minimal expected costs under $F_x$
\begin{equation}\label{eq:opt_action}
    \delta_c(F_x) \coloneqq \underset{a \in \mathcal{A}}{\arg\min} \; \mathbb{E}_{Y \sim F_x} [c(a, Y)].
\end{equation}
When the decision maker trusts the forecast, then $\delta_c(F_x)$ is the optimal action.\footnote{Here, "trust" means that the decision maker assumes the true observations $Y$ are distributed according to $F_x$.}
Here, the expected value of the cost function translates the abstract concept of uncertainty into an interpretable and actionable quantity: the expected cost of a particular action. The expectation automatically accounts for the fact that even low probability events can have a high impact if they are costly.
The observed costs are then given by 
\begin{equation}\label{eq:obs_costs}
    C_{obs} = \mathbb{E}_{X,Y} [c(\delta_c(F_X), Y)]    
\end{equation}
and measure the actual costs when taking action $\delta_c(F_X)$ and then observing the true values of $Y$. Good forecasts lead to good decisions and thus yield low observed costs.
Different cost functions place emphasis on different regions of the outcome space $\mathcal{Y}$. In order to enable good decisions, forecasts must be well calibrated in those parts of $\mathcal{Y}$ that are most relevant for the decision problem at hand.
Consider a binary decision task, $\mathcal{A} = \{a_0,a_1\}$. On some subsets $\tilde{\mathcal{Y}} \subset \mathcal{Y}$, decisions may be trivial, e.g., when $c(a_0,y) < c(a_1,y) \; \text{for all } y \in \tilde{\mathcal{Y}}$ and calibration is unnecessary. However, in regions where $c(a_0, \cdot)$ and $c(a_1, \cdot)$ intersect, good calibration is essential for correctly estimating expected costs and making optimal decisions. Thus, cost functions encode costs and preferences specific to a decision maker, enforcing calibration only where it matters for the decision task.

\subsection{Relation to proper scoring rules}
\label{sec:psr_connection}
At first glance, evaluating forecasts at the decision level seems very different from traditional forecast verification metrics such as the CRPS or other proper scoring rules. However, both approaches are closely linked through the framework of cost-induced scoring rules \citep{hartline2025}. For a forecast $F_x$ and a corresponding observation $y$, every bounded cost function $c$ induces a proper scoring rule $S_c$ via the best-response cost 
\begin{equation*}
    S_c(F_x, y) = c(\delta_c(F_x), y)\, ,
\end{equation*}
that is, the cost incurred by a decision maker who chooses the optimal action based on the forecast $F_x$ \citep{hu2024, gneiting2007strictly}. Conversely, for every proper scoring rule $S$, there exists a cost function $c$ that induces $S$. Since the CRPS is a (strictly) proper scoring rule, it is induced by some cost function. Thus, choosing the CRPS for forecast verification implicitly assumes that the underlying cost structure is appropriate for the task at hand. While this cost structure may reward good overall calibration, it may place insufficient weight on those parts of the predictive distribution that are most relevant for a specific downstream decision task. Hence, knowledge of the decision task provides a principled basis for selecting an appropriate cost-induced scoring rule for forecast evaluation.

\subsection{Decision calibration}
Decision calibration \citep{zhao2021} holds when the expected costs (of the optimal action) under the forecast match with the true observed costs. The cost gap is the absolute difference between the two and a measure of decision (mis-) calibration. More precisely, the expected cost of the optimal action under the forecast is given by
\begin{equation}\label{eq:exp_costs}
    C_{exp} = \mathbb{E}_X \mathbb{E}_{\hat{Y} \sim F_X} [c(\delta_c(F_X), \hat{Y})]
\end{equation}
and can be seen as a simulation of the incurred costs.
The cost gap is then given by
\begin{equation}
    C_{gap} = |C_{exp} - C_{obs}| \,,
\end{equation}
where $C_{obs}$ is defined as in \Cref{eq:obs_costs}. A smaller cost gap indicates better decision calibration, with perfect calibration achieved for $C_{gap} = 0$.\footnote{The expectation over $X$ does not need to be taken over the whole domain $\mathcal{X}$ but instead can be over subsets of $\mathcal{X}$ or in the extreme case individual $x \in \mathcal{X}$, leading to stronger conditions than the mere equivalence of average expected and true costs.}
In a good forecast, both the cost gap and the observed cost should be small: forecasts should correctly anticipate the incurred costs but also lead to small incurred costs, i.e., good decisions.

\section{Weather forecasts and decision-making}
\label{sec:decision_making}
Weather forecasts are used for decision-making, so it seems natural to evaluate their performance at the decision level rather than the forecast level. Weather forecasting is therefore predestined for a decision calibration analysis. In the following, we introduce and motivate weather-dependent decision tasks and explain how decision calibration is evaluated.

\subsection{Decision tasks}
There are many decision tasks related to weather, yet we were unable to find any that have been explicitly formulated as cost functions.\footnote{Real-world cost functions may often be commercially or politically sensitive and therefore not publicly available.} Therefore, we define exemplary cost functions for decision tasks related to temperature and wind speed forecasts. 
While these cost functions are motivated by plausible real-world applications, our goal is not to recover the true cost function of any particular decision maker. Doing so would require detailed, context-specific knowledge of their costs and preferences.
Rather, we use parameterized cost functions to represent classes of plausible decision makers to investigate the interplay between forecast calibration and decision calibration. The resulting cost functions are then used to simulate a Bayes decision-making process.
We emphasize that in concrete applied settings, when sufficient context-specific knowledge is available, such cost functions can in principle be derived using established procedures from applied economics \citep{schlenker2009, kamen2026} and, in particular, insurance \citep{kumar2026}. Additionally, insights from cognitive and social science research on weather-related decision making (e.g., \citealt{joslyn2012}) can inform cost function design for individual decision-makers. This supports the deployment of our forecast evaluation method by practitioners in highly context-specific settings.

\begin{figure}
    \centering
    \includegraphics[width=0.8\linewidth]{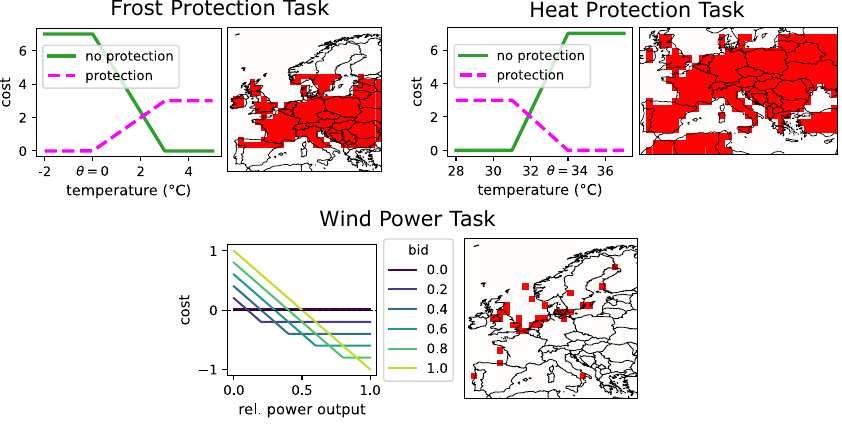}
    \caption{Cost functions and evaluated grid points (red) for the frost protection task, heat protection task (cost ratio, $r=0.7$), and wind power task (shortfall penalty, u-pen = 2). For visual clarity, we only plot costs for every other action (bid) in the wind power task.}
    \label{fig:cost_fns}
\end{figure}

\paragraph{Frost protection}
In the US, frost causes the biggest economic damage among all meteorological events \citep{juurakko2021,white1975}.
We motivate the cost function with the example of plant frost protection in the agricultural sector. However, our cost function could also model frost decision processes in other weather-dependent domains, such as infrastructure, e.g, road anti-icing deployment decisions.
In the agricultural sector, farmers need to decide whether to employ frost protection measures (covering plants, active heating, or early harvesting) when freezing temperatures are likely \cite{snyder2005}. When temperatures drop too low and no protective measures are in place, plants die, causing crop and economic losses. However, these protective measures are costly and should only be applied when necessary.
We model a binary decision task, whether to apply frost protection, defining a threshold temperature $\theta \in [-4,4]~^\circ$C below which the plants die ($t_{obs} \leq \theta$) and consider them safe if $t_{obs} \geq \theta +3$. The range $[-4, 4]~^\circ$C is centred symmetrically around 0°C, covering mild to moderately severe frost thresholds relevant for a variety of frost-sensitive applications in European climates. The 3°C buffer reflects the gradual onset of frost damage as temperatures approach the critical threshold from above, consistent with continuous damage response curves used in frost models \citep{gabbrielli2022review, salazar2016freezing}.\footnote{The exact width of this transition zone is context-dependent, varying for instance across crop species. Results were found to be robust to variations in this buffer size.}
Without protection, we incur the maximum cost (all crops are lost) when plants die, and no cost when they are safe.
For intermediate temperatures, we interpolate linearly. With protection, we flip the cost function: incurring a cost when protection is unnecessary and no cost when it saves the crop. We use asymmetric costs, where the cost ratio $r \in (0,1)$ weights costs of missed protection ($r \cdot 10$) and unnecessary protection $(1-r) \cdot 10$ (see \Cref{fig:cost_fns} for an example configuration).\footnote{We scale costs by a factor of 10 following \citet{sahoo2021}; results are invariant to this scaling.} By varying $\theta$ and $r$, we can model decision makers cultivating plants with different temperature sensitivities and protection costs. We evaluate performance in Europe for 00 UTC forecasts, considering mid-latitudes where agriculture and frost are likely.

\paragraph{Heat protection}
Extreme weather events like heavy rain or snowfall, floods, and heatwaves can pose a serious threat to the population. Since the frequency of extreme events is expected to increase due to climate change, reliable early warning systems become increasingly relevant \citep{eriksen2023}. Heat events are responsible for high numbers of heat-related illnesses and deaths every year. 
We motivate the cost function with the example of civil heat protection.
Heat-related deaths can be prevented, for instance, by protecting vulnerable groups, but effective protective measures require early warning systems \citep{luber2008, meerow2022, hess2023} and come at a cost, so they should only be implemented when required.
We model a binary decision task, whether to apply heat protection, using the same cost function design as in the frost protection task above, but with $\theta$ now marking the temperature of maximum heat-related cost (see \Cref{fig:cost_fns}). For larger $\theta$, the decision task will often be moved into the tails of the distribution. Again, we vary the threshold temperature $\theta \in [30,38]~^\circ$C 
and cost ratio $r$ to model multiple decision makers (heat sensitivities and costs) and evaluate performance in Europe for 12 UTC forecasts, excluding Scandinavia. The range of 30--38°C covers heat stress thresholds relevant across European climates, from temperate regions where 30°C already constitutes a heat stress event to warmer southern European climates where thresholds are higher \citep{casanueva2019overview}.

\paragraph{Wind power dispatch}

The availability of renewable energy sources, such as wind and solar, depends on the weather.
A large-scale expansion of renewable energy, therefore, requires forecasts that can reliably predict their availability so that dispatch plans can be made to ensure power system stability \citep{botterud2011, sweeney2020}. In the case of wind power, this leads to a dispatch decision task, where a wind power provider needs to decide in advance how much power he will provide at some point in the future and deviations from the guaranteed amount can lead to penalties \cite{bourry2008}. We model this as a multi-action decision task, where each action corresponds to a particular amount of relative power that the provider promises to deliver (see \Cref{fig:cost_fns}). We consider 11 actions where 10 actions partition the relative power space into equal-sized bins ($0.1,0.2,...,1.0$). The more energy the provider promises, the more he gets paid. However, in the case of a production deficit, he needs to cover the shortfall by buying (expensive) energy from other sources, which increases his costs (linear cost penalty).
In the case of a production surplus, wind farms can reduce the power output (curtailment) to not exceed the target amount, hence, no penalty costs apply \citep{bruninx2025}. In this case, however, the provider would have been better off by predicting a higher power output. We model such opportunity costs implicitly through the (lower) cost of other actions. Additionally, there is a 'turbine-off' action with a very small constant cost (standby drain) when no power is expected due to low or too high wind speeds. We evaluate performance on grid points in Europe with offshore wind farms\footnote{Dataset on offshore wind farms provided by the European Marine Observation and Data Network (EMODnet).}for 00 UTC forecasts. We vary the penalty for under-delivering (u-pen $\in \{2,3,4\}$) by changing the slope of the cost function. The considered penalty range covers realistic imbalance penalty ratios and should be viewed as conservative relative to the imbalance price spikes observed in European balancing markets, for example the German market \citep{narajewski2022probabilistic}. We use the power law to extrapolate wind speeds from 10 meters to turbine hub height, then convert to wind power using a standard power curve; for details, see \Cref{sec:appdx:windpower_task}.

\subsection{Weather prediction models}
We consider a standard numerical weather prediction model and a novel machine learning weather prediction model.
The IFS ENS is the probabilistic NWP model operational at the European Center for Medium-Range Weather Forecasts (ECMWF) that provides ensemble forecasts. Each ensemble member is generated by running the numerical solver with slightly perturbed initial conditions and physical parameters, leading to different forecasts.
For the MLWP model, we use ArchesweatherGen \citep{couairon2024}, a generative ML weather prediction model that provides probabilistic weather forecasts. We will refer to it as Arches from now on. Arches is based on a diffusion model \citep{ho2020} trained to learn the true (conditional) data distribution from historical reanalysis data (ERA5, \citep{hersbach2020}). Once trained, the model generates samples from the learned forecast distribution (for a 24 h lead time) at a horizontal resolution of $1.5^{\circ}$ given the two previous atmospheric states. Forecasts with longer lead times are obtained through autoregressive rollouts.

\subsection{Evaluating decision calibration of weather forecasts}
In weather forecasting, we input the state of the atmosphere at time $t$, $X_{t}$, and aim to predict the state of the atmosphere at lead time $l$, $X_{t+l}$. Let $Y_{t+l}$ denote the weather variable of interest at a given grid point (e.g., 2 meter temperature) extracted from $X_{t+l}$. We evaluate ensemble forecast models, that provide $M$ samples from the forecast distribution $F^l_{X_t}$
\begin{equation}
    \{\hat{Y}^{(i)}_{t+l}\}_{i=1}^M \sim F^l_{X_t} \,.
\end{equation}
We approximate the expected costs in Eq. (\ref{eq:opt_action}) via a Monte Carlo estimate 
\begin{equation}
    \mathbb{E}_{\hat{Y}_{t+l} \sim F_x^l} [c(a, \hat{Y}_{t+l})] \approx \frac{1}{M} \sum_{i=1}^M c(a,\hat{Y}^{(i)}_{t+l}) \,.
\end{equation}
We consider single surface variables, hence for every lead time, grid point, and day of the year, we obtain an ensemble forecast and a corresponding observation. Given a particular decision task, we compute the expected cost for every ensemble forecast, select the optimal action and compare expected cost to observed cost using the corresponding observation to obtain the cost gap. That is, we compute the cost gap for every observation, rather than first averaging costs across observations as in \Cref{eq:exp_costs}. We do this because average cost comparisons can mask poor performance of individual forecast instances, as discussed in \cite{zhao2021}. The cost gap depends on both errors in the forecast distribution and the structure of the cost function. After computing the cost gap and observed cost for every observation, we compute averages of both quantities over space and time.
\section{Experiments}
We consider 50 member ensemble forecasts from IFS ENS and Arches for the year 2021 for 2 meter temperature and 10 meter windspeed with lead times up to 15 days. We coarsen the horizontal resolution of IFS to $1.5^{\circ}$ to match that of Arches. Following \cite{price2023}, we use the IFS analysis (IFS HRES fc0) and ERA5 \citep{hersbach2020} as ground-truth observations for IFS ENS and Arches, respectively. All data is available in the Weatherbench dataset \citep{rasp2024}, except for Arches forecasts, which are obtained by running the trained public model. For each task we compare standard calibration results (see Appendix \Cref{sec:appdx:cal} for an overview) to decision calibration results.

\begin{figure}[H]
    \centering
    \includegraphics[width=0.77\linewidth]{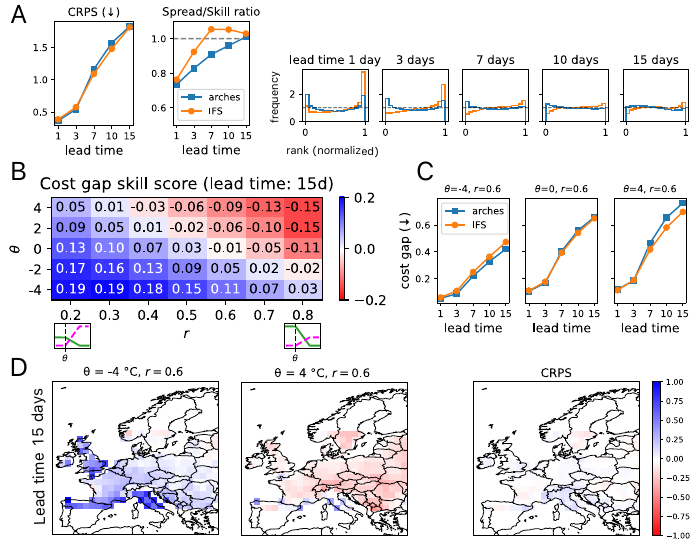}
    \caption{Frost protection task. (A) Results for standard calibration analysis (2 meter temperature) indicate similar performance between both models (particularly for 15 days lead time). (B-D) Results for decision calibration analysis show how cost gap improvements vary across tasks. (B) and (D) show relative improvements (Arches over IFS) at 15 days lead time, where a value of $0.1$ indicates a $10 \%$ improvement. (C) shows absolute cost gaps over lead time for selected task parameters ($\theta \in \{-4, 0, 4\}$°C, $r=0.6$).}
    \label{fig:frost_results}
\end{figure}

\subsection{Frost protection}
CRPS results for 2 meter temperature (00 UTC) indicate similar performance between both models, with Arches performing slightly worse for 7 and 10 days lead time (see \Cref{fig:frost_results}). SSR results indicate a stronger underdispersion for Arches and some overdispersion for IFS for longer lead times. The PIT histograms indicate similar performance for lead times longer than 3 days. However, when we consider the frost protection task results (\Cref{fig:frost_results} B-D), we can see that decision calibration (cost gap) improvements of Arches over IFS ENS vary across tasks. For low threshold temperatures ($\theta$), Arches yields a better decision calibration than IFS (up to $19\%$ smaller cost gap compared to IFS at 15 days lead time), but as $\theta$ increases, this effect weakens or even reverses (up to $15\%$ larger cost gap compared to IFS at 15 days lead time). 
We observe a similar effect for the cost ratio $r$. This makes sense as small (large) values of $r$ effectively move the decision boundary of the task further towards lower (higher) temperatures. High values of $r$ correspond to settings where protection costs are small relative to potential crop losses, while $\theta$ controls frost sensitivity of the crop: high $\theta$ represents frost-sensitive crops (damaged at mild cold), while low $\theta$ represents hardy crops (only damaged at severe cold).
Note that directional biases can coincidentally benefit certain decision tasks when aligned with the cost asymmetry (see Appendix \Cref{sec:appdx:bias} for a detailed discussion). However, PIT histograms show no systematic bias at the lead times where the largest differences in decision calibration are observed (10 -- 15 days). Furthermore, differences in decision calibration persist under symmetric costs ($r=0.5$), where cost asymmetry cannot cause one bias direction to be more beneficial than another. This suggests that systematic bias is not a primary source of the observed differences in decision calibration.
Consistent with the cost gap results, we see a similar trend in observed costs: improvements in decision calibration generally lead to lower costs (see \Cref{fig:appdx:frost_task:obs_cost}). At a lead time of 15 days, choosing Arches over IFS yields up to $15\%$ lower costs for low threshold temperatures, but up to $11\%$ higher costs for larger threshold temperatures. This result highlights the importance of task-dependent evaluations - even small changes in the task can lead to different conclusions about model performance. An effect that goes unnoticed with global calibration metrics.

\subsection{Heat protection}

\begin{figure}[H]
    \centering
    \includegraphics[width=0.77\linewidth]{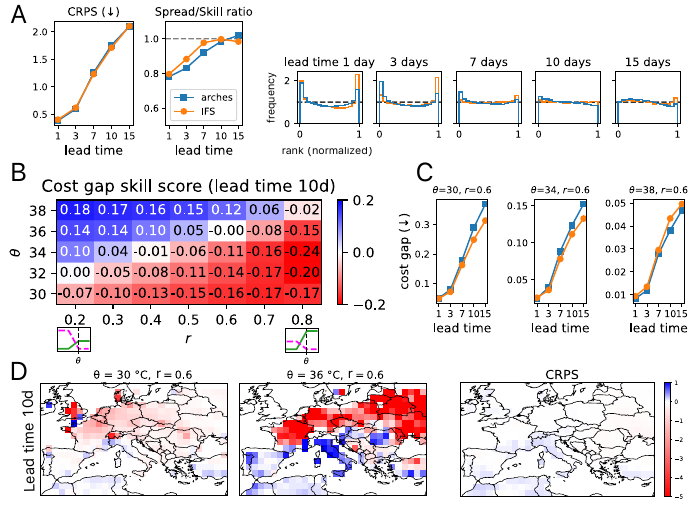}
    \caption{Heat protection task. (A) Results for standard calibration analysis (for 2 meter temperature) indicate similar performance between both models. (B-D) Results for decision calibration analysis show that cost gap improvements vary across tasks. (B) and (D) show relative improvements (Arches over IFS) at 10 days lead time, where a value of $0.1$ indicates a $10 \%$ improvement.
    (C) shows absolute cost gaps over lead time for selected task parameters ($\theta \in \{30, 34, 38\}$°C, $r=0.6$).}
    \label{fig:heat_results}
\end{figure}

Both CRPS and PIT histogram results indicate similar calibration performance of both models for 2 meter temperature (12 UTC). SSR results indicate a slightly stronger underdispersion for Arches (see \Cref{fig:heat_results}).
However, for the decision task, we observe a similar pattern as in the frost protection task: different tasks lead to different conclusions about model performance (see \Cref{fig:heat_results} B - D). Low values of $\theta$ with high $r$ could represent settings such as urban public health agencies issuing heat warnings in temperate cities, where populations are less heat-adapted and warning costs are low relative to health impacts. Decision calibration in Arches improves with increasing temperature, often surpassing that of IFS ENS at high temperatures ($\theta$). Following the same reasoning as for the frost protection task, systematic bias is an unlikely driver of the observed differences in decision calibration. We see a similar (but less pronounced) pattern for the observed costs (see \Cref{fig:appdx:heat_task:obs_cost}). At 10 days lead time and high temperature thresholds, Arches yields up to $18\%$ smaller cost gap and $9\%$ lower costs than IFS. By plotting decision calibration improvement per grid point, we see that model rankings also vary across geographical regions - an effect that can be lost through the aggregation over space. In the Mediterranean, Arches performs better than in central Europe; this effect is stronger at higher temperature thresholds.

\subsection{Wind power dispatch}

\begin{figure}[H]
    \centering
    \includegraphics[width=0.75\linewidth]{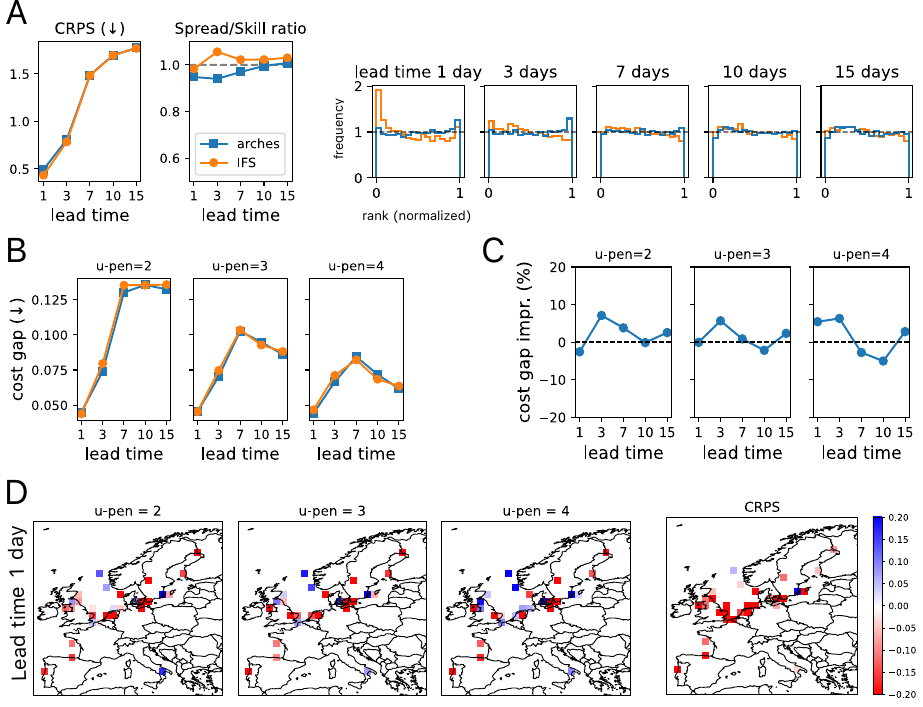}
    \caption{Wind power dispatch task. (A) Results for standard calibration analysis and (B-D) decision calibration analysis. (C) and (D) show relative improvements of the cost gap (Arches over IFS). (D) shows relative improvement per wind farm location (nearest grid point) at 1 day lead time, where a value of 0.1 indicates a 10\% improvement.}
    \label{fig:windpower_results}
\end{figure}

We report standard calibration results for wind speeds.\footnote{We consider the wind power transformation to be part of the downstream task.} Both CRPS and PIT histogram suggest very similar performance between Arches and IFS (\Cref{fig:windpower_results} A) except for 1 day lead time, where both diagnostics disagree: IFS attains a better CRPS score but a worse PIT histogram that suggests a slight overestimation of wind speeds. SSR results indicate a slight underdispersion for Arches and overdispersion for IFS. The decision calibration analysis reveals only small cost gap differences between both models that vary in strength and direction across lead times (see \Cref{fig:windpower_results} B-D). Additionally, we find only small differences in the observed costs, with Arches performing slightly better than IFS (see \Cref{fig:appdx:windpower_task:obs_cost}). As penalties increase, cost gaps decrease for longer lead times. This is because, for long lead times, we observe larger forecast uncertainties, so the 'shut-off' action often has the lowest expected cost, and, since it has constant cost, the cost gap is trivially 0. For some lead times, we obtain opposite rankings of the cost gap and the observed costs. We conjecture that this is due to aggregation effects (over multiple actions) and limited sample size. 
Wind power trading happens mostly on the day-ahead market, so 1 day lead time performance is most relevant for this task. At this lead time, we see that for CRPS and SSR, Arches performs slightly worse than IFS. Yet, for decision calibration, Arches outperforms IFS with stronger penalties, yielding a lower cost gap (see \Cref{fig:windpower_results} C, u-pen=4, lead time 1 day). This may be partly explained by the slight overestimation of wind speeds by IFS at this lead time (cf. PIT histogram, \Cref{fig:windpower_results} A), which would systematically increase costs under larger shortfall penalties.
Note that \Cref{fig:windpower_results} D shows per-wind-farm relative improvements at 1 day lead time, which can look more IFS-favoured than the aggregate cost gap comparison in panel C, as small cost gap increases over IFS can appear as large relative deficits (red) when the IFS baseline is itself small.\footnote{Panel D shows relative improvements per wind farm location, while panel C first aggregates cost gaps before taking the relative comparison. At locations where IFS's cost gap is small, a small worsening by Arches translates into a large relative deficit (red) without contributing much to the aggregate, so red-heavy panels can coexist with an Arches-favoured aggregate.}
Arches yields slightly lower observed costs than IFS (see \Cref{fig:appdx:windpower_task:obs_cost} B). 
Although the differences are small, the absolute savings can still be meaningful at high revenue volumes.
We conjecture that the small differences in cost gap rankings are due to the fact that the particular cost function in this example requires decisions across the whole output domain, and errors are punished similarly for all actions. In this case, decision calibration essentially probes calibration at the global level (rather than on a subset of the domain), and thus model rankings deviate less from those obtained through standard verification metrics, such as CRPS.

\section{Conclusion}
\label{sec:conclusion}
In this work, we apply the novel decision calibration framework to evaluate weather forecasts.
Our results show that performance on typical calibration metrics does not reliably translate into decision-making performance.
In a frost protection task, model rankings for decision calibration vary across seemingly similar decision tasks.
In a heat protection task, differences between models emerge only when evaluating calibration at the decision level, demonstrating that decision calibration can elegantly assess forecast calibration for tail events. 
In a wind power dispatch task, we only observe small differences in decision calibration and observed costs between the two models.
We conjecture that decision calibration rankings deviate less from global calibration rankings when cost functions require decisions across the entire output space and errors are punished similarly across actions.
One limitation of decision calibration is the requirement of well-defined cost functions, which may not be readily available in practice.
However, as discussed in \Cref{sec:decision_making}, established methods from applied economics and cognitive science provide a practical path to deriving such cost functions in concrete applied settings. Our results highlight the need to evaluate forecast performance at the decision level to identify the best model for a particular decision task. Decision calibration, therefore, offers a promising direction for tailored, task-specific probing of forecast performance. 
While our evaluation focuses on weather forecasting, the framework applies broadly to any domain where probabilistic models inform downstream decisions, where standard calibration diagnostics may similarly fail to predict decision performance.
Future work could optimize ML weather prediction models for specific downstream decision tasks by utilizing cost functions as the loss function during training or fine-tuning. Because cost functions induce proper scoring rules, such as the CRPS commonly used in state-of-the-art probabilistic ML models, this would constitute a natural extension of current training practice and provide a principled way to align model optimization directly with the relevant decision task.

\bibliographystyle{apalike}  
\bibliography{references}  

\newpage
\appendix
\section*{Appendix}
\renewcommand{\thefigure}{\thesection.\arabic{figure}}
\section{Decision Tasks}
\setcounter{figure}{0}

\subsection{Wind power dispatch} \label{sec:appdx:windpower_task}
\begin{figure}[H]
    \centering
    \includegraphics[width=0.3\linewidth]{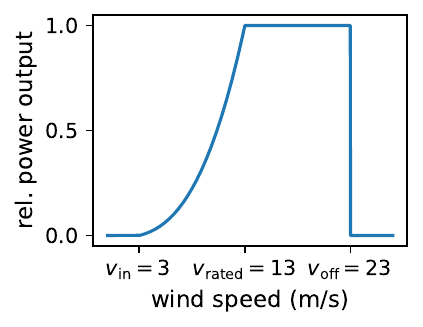}
    \caption{Wind power curve used to convert wind speed into wind power.}
    \label{fig:appdx:wind_task:powercurve}
\end{figure}

We extrapolate 10 meter wind speeds ($v_{10}$) to turbine hub height ($h$) wind speeds ($v_h$) using the power law \citep{abbes2012}
$$v_h = v_{10} \cdot (\frac{h}{10})^\alpha \,,$$ with an exponent $\alpha = 0.1$ (accounting for low offshore surface roughness) and hub height $h = 120$m.
We further transform wind speeds into wind power using a standard wind power curve (\Cref{fig:appdx:wind_task:powercurve}) using a cubic polynomial curve modeling approach \citep{wang2019}. We set the cut-in wind speed ($v_{in}$) to 3 m/s, the rated wind speed ($v_{rated}$) to 13 m/s and the cut-off wind speed ($v_{off}$) to 23 m/s.

\section{Calibration diagnostics in weather forecasting} \label{sec:appdx:cal}
Calibration of weather forecasts is typically verified using the following diagnostic tools. PIT histograms \citep{gneiting2007} are used to assess the (rather weak) notion of probabilistic calibration. The probability integral transform (PIT) is obtained by evaluating the predictive CDF at the observation and probabilistic calibration holds when the PIT values follow a uniform distribution. However, probabilistic calibration is rather weak as it only measures forecast calibration on average (over all forecasts) and so no statements can be made about the calibration of individual forecast or even subsets of forecasts. The continuous ranked probability score (CRPS) \citep{gneiting2007} is a (strictly) proper scoring rule that quantifies the quality of a probabilistic forecast via a numerical score. Strictly proper scoring rules consider both calibration and sharpness of a forecast and are uniquely minimized if and only if the forecast distribution aligns with the true data generating distribution. The CRPS provides a stronger diagnostic tool than the PIT histogram and is widely employed in the assessment of probabilistic weather forecasts. Finally, the spread-skill ratio (SSR) checks the assumption of exchangeability of the observation and the ensemble members of the forecast to assess calibration \citep{fortin2014}. It is given by the ratio of the average ensemble spread and the ensemble skill (RMSE of ensemble mean). Intuitively, forecast spread should be a predictor of the skill of the ensemble mean. An SSR of $1.0$ is ideal with smaller (larger) values corresponding to underdispersed (overdispersed) ensemble forecasts.\footnote{Strictly speaking, an SSR of 1.0 indicates slight over-dispersion, as observation errors inflate the RMSE component but are not represented in the ensemble spread. To correct this, one would need to subtract the effect of observation errors on the RMSE. However, as observation error estimates are typically unavailable, the magnitude of this effect cannot be assessed and the correction is rarely applied in practice.} Similar to the PIT histogram, the SSR compares spread and skill marginally so conclusions about individual forecast performance can not be drawn. It is important to highlight that both PIT histogram flatness as well as SSR values close to 1.0 are only necessary but insufficient conditions for well-calibrated forecasts, in the sense of distribution calibration \citep{dirkson2025}. For probabilities of specific binary events, additional verification metrics become applicable. These include the Brier score and its decomposition into reliability and resolution \citep{murphy1973}, reliability diagrams, and the Relative Operating Characteristic (ROC) curve \citep{mason2002}. The ROC curve captures a forecast system's ability to discriminate between the occurrence and non-occurrence of the event across all probability thresholds. Reliability and resolution measure, respectively, whether forecast probabilities match observed event frequencies, and the extent to which forecast probabilities vary with the actual outcome. As our focus is on the full predictive distribution rather than specific binary events, we rely on the distribution-level diagnostics described above (CRPS, SSR, PIT), which are the standard verification suite in current ML-based forecast systems.

\section{Experiments}
This section presents additional results for all three decision tasks discussed in the main paper.
While the main paper focuses on the cost gap $C_\text{gap}$ (Eq.~(4)) as the primary decision calibration diagnostic, most of the figures here show observed costs $C_\text{obs}$ (Eq.~(2)), providing an additional view of model performance.
\setcounter{figure}{0}
\subsection{Frost protection}
\begin{figure}[H]
    \centering
    \includegraphics[width=0.77\linewidth]{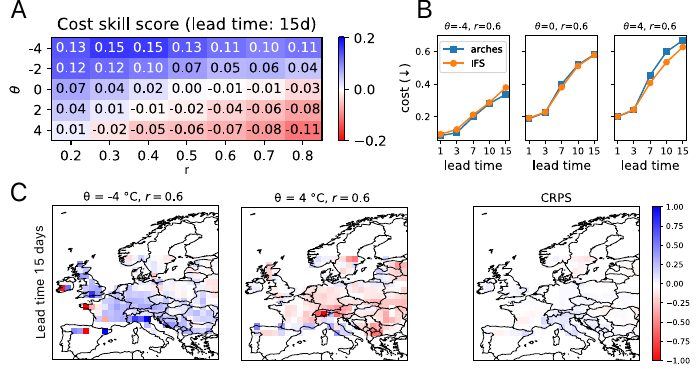}
    \caption{Frost protection task: Same as \Cref{fig:frost_results} but for observed costs rather than cost gaps.
    (A) and (C) show relative improvements of Arches over IFS at 15 days lead time (0.1 means 10\% improvement).}
    \label{fig:appdx:frost_task:obs_cost}
\end{figure}

\Cref{fig:appdx:frost_task:obs_cost} shows observed costs for the frost protection task.

\subsection{Heat protection}
\begin{figure}[H]
    \centering
    \includegraphics[width=0.87\linewidth]{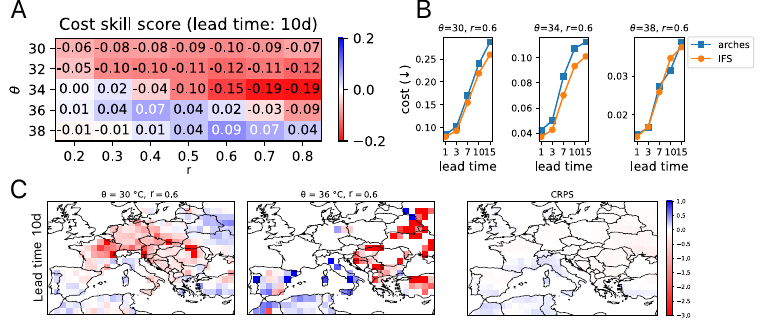}
    \caption{Heat protection task: Same as \Cref{fig:heat_results} but for observed costs rather than cost gaps. (B) shows absolute costs (Arches in blue, IFS in orange). (A) and (C) show relative improvements of Arches over IFS at 10 days lead time (0.1 means 10\% improvement). (C) For $\theta = 36$ °C many grid points in central Europe show no differences between both models. At those grid points, temperatures are not high enough; hence both models lead to the 'no protection' action and obtain the same costs.}
    \label{fig:appdx:heat_task:obs_cost}
\end{figure}

\Cref{fig:appdx:heat_task:obs_cost} shows observed costs for the heat protection task.

\subsection{Wind power dispatch}
\label{sec:appdx:exp:windpower_task}
\begin{figure}[H]
    \centering
    \includegraphics[width=0.7\linewidth]{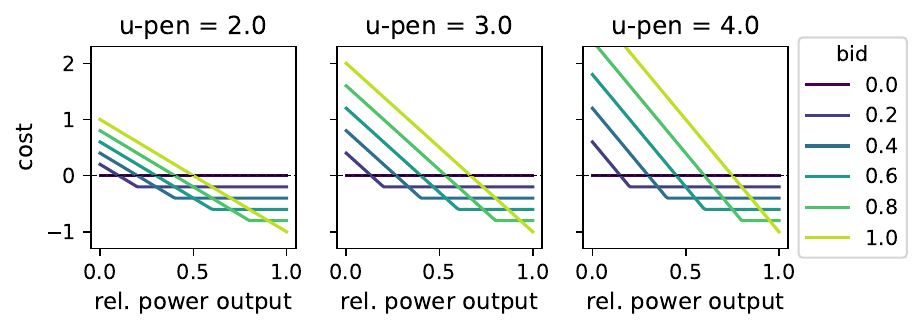}
    \caption{Wind power dispatch task: cost functions for shortfall penalties u-pen $\in \{2, 3, 4\}$ from left to right.}
    \label{fig:windpower_cost_fns}
\end{figure}

\begin{figure}[H]
    \centering
    \includegraphics[width=0.7\linewidth]{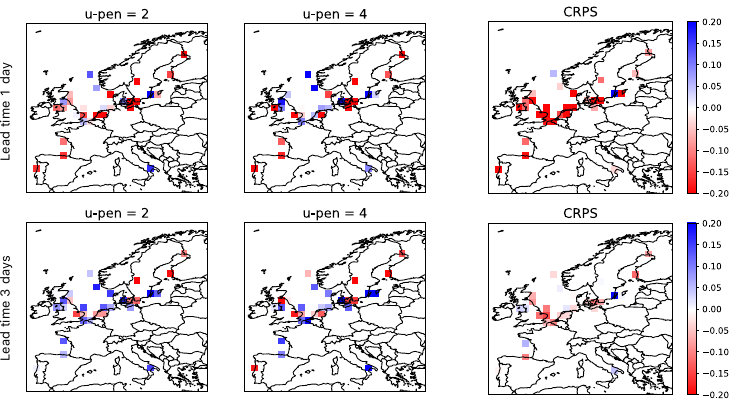}
    \caption{Wind power dispatch task: relative improvements of cost gap and CRPS (Arches over IFS) per wind farm location. At 1 day lead time, we see improvements in decision calibration for larger penalties (u-pen = 4). For 3 days lead time, differences to the CRPS ranking are more pronounced and similar across tasks (shortfall penalties).}
    \label{fig:appdx:windpower_task:map}
\end{figure}

\begin{figure}[H]
    \centering
    \includegraphics[width=0.77\linewidth]{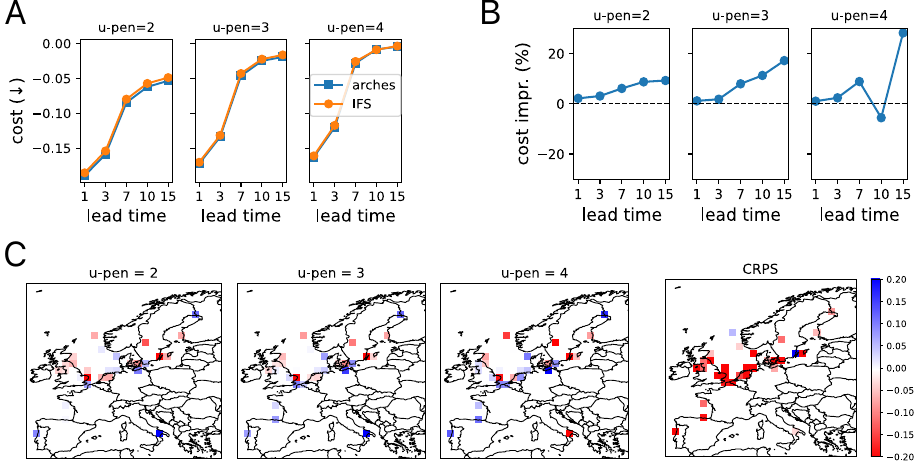}
    \caption{Wind power dispatch task: Same as \Cref{fig:windpower_results} but for observed costs rather than cost gaps. (A) We see slightly lower observed costs for Arches across lead times and tasks. (B) Relative improvements of observed costs (Arches over IFS). (C) Relative improvements per wind farm location for observed costs and CRPS at 1 day lead time (0.1 means $10\%$ improvement).}
    \label{fig:appdx:windpower_task:obs_cost}
\end{figure}

\Cref{fig:windpower_cost_fns} shows the cost functions for three shortfall penalties. \Cref{fig:appdx:windpower_task:map} extends the spatial results from the main paper (\Cref{fig:windpower_results} D) showing both 1 and 3 days lead time. \Cref{fig:appdx:windpower_task:obs_cost} shows observed costs for the wind power dispatch task, complementing \Cref{fig:windpower_results} of the main paper which shows cost gaps.

\section{Directional biases and decision calibration}
\label{sec:appdx:bias}
Systematic biases in probabilistic forecasts can influence decision calibration results. For decision tasks with a threshold structure in the cost function, bias direction affects the balance of false positive and false negative decisions. With asymmetric costs, one error type is more costly than the other.
A bias that increases the frequency of the costly error type, for example, missed frost events in a frost protection task, will therefore lead to higher observed costs and a larger cost gap. On the other hand, a bias that increases the frequency of unnecessary protection (a less costly error) will have a smaller impact and yield lower observed costs than the opposite bias direction. A bias that increases costly errors also inflates the cost gap (discrepancy between anticipated and observed costs), since the potential gap is larger when a costly error occurs.
The exact magnitude and direction of this effect depend on the data distribution around the decision boundary and are therefore task-specific. Note that no biased forecast can yield lower observed costs than the unbiased forecast in expectation, as observed costs constitute a proper scoring rule (see \Cref{sec:psr_connection}).

Our framework captures both bias-driven and distributional sources of suboptimal decision quality. A model that is biased in the costly direction will score poorly, as will a model that is poorly calibrated near the decision boundary, since both ultimately harm downstream decisions.

\end{document}